\documentclass[a4paper,%
  12pt,%
  abstracton,%
  footexclude,%
  normalheadings,%
  pointednumbers,%
  halfparskip,%
]{scrartcl}

\newcommand{\UPLB}{University of the Philippines Los Ba\~{n}os}

\begin{document}

\title{The Interactive Effects of Operators and Parameters to GA Performance Under Different Problem Sizes}
\author{
   Jaderick P. Pabico and Elizer A. Albacea\\
   \small{Institute of Computer Science}\\
   \small{\UPLB}\\
}
\date{}
\maketitle

\begin{abstract}
The complex effect of genetic algorithm's (GA) operators and parameters to its performance has been studied extensively by researchers in the past but none studied their interactive effects while the GA is under different problem sizes. In this paper, We present the use of experimental model (1)~to investigate whether the genetic operators and their parameters interact to affect the offline performance of GA, (2)~to find what combination of genetic operators and parameter settings will provide the optimum performance for GA, and (3)~to investigate whether these operator-parameter combination is dependent on the problem size. We designed a GA to optimize a family of traveling salesman problems (TSP), with their optimal solutions known for convenient benchmarking. Our GA was set to use different algorithms in simulating selection ($\Omega_s$), different algorithms ($\Omega_c$) and parameters ($p_c$) in simulating crossover, and different parameters ($p_m$) in simulating mutation. We used several $n$-city TSPs ($n=\{5, 7, 10, 100, 1000\}$) to represent the different problem sizes (i.e., size of the resulting search space as represented by GA schemata). Using analysis of variance of 3-factor factorial experiments, we found out that GA performance is affected by $\Omega_s$ at small problem size (5-city TSP) where the algorithm Partially Matched Crossover significantly outperforms Cycle Crossover at $95\%$ confidence level. Under intermediate problem sizes (7-city and 10-city TSPs), we found out that the mean GA performance is affected by the $\Omega_s \times \Omega_c$ interaction where the average performance of GA across $p_c$ and $p_m$ varies at different $\Omega_s$-$\Omega_c$ combinations. At big problem sizes (100-city and 1000-city TSPs), we observed that a 3-way interaction among $\Omega_s$, $\Omega_c$, and $p_m$ exist to affect the GA performance averaged across different $p_c$. Similarly, we also observed that the  3-way interaction among $\Omega_s$, $p_c$ and $p_m$ affects the GA performance averaged across all $\Omega_c$. To explain these three-way interactions, we used the Duncan's Multiple Range Test at $5\%$ probability level to perform pairwise comparison of means of GA performance. 
\end{abstract}


\section[Introduction]{Introduction}

Genetic Algorithms (GAs) are probabilistic search techniques suited for solving large, complex, multidimensional, multimodal, discontinuous, and/or noisy search and optimization problems. Applied to such problems, GAs outperformed several tested search and optimization procedures such as the gradient techniques and some various forms of random search~\cite{Beaty93,Davis87b,Davis91,Goldberg89,Grefenstette89,Holland92}. In the past years, the GA algorithms for selection, crossover, and mutation and the GA parameters population size, crossover probability, and mutation rate have received much attention in research~\cite{Lee94,Back92,Schraudolph92}. These studies show that depending on the operators used and the parameter setting, the behavior of the GA can range from that of random search to hill climbing~\cite{Lee94}. Thus, designing a GA that would meet a specific problem domain's resource constraints would require a significant effort in trying to find out the right GA operator-parameter combination. 

Many researchers have attempted to find a set of genetic operators and parameters for GAs to perform optimally for solving a given problem domain~\cite{DeJong75,Grefenstette86,Schaffer89,Davis91,Back92,Schraudolph92,Lee94}. These researchers have used techniques such as hand optimization, a meta-GA, brute force search, and adapting parameters which are costly and time consuming~\cite{DeJong75,Grefenstette86,Schaffer89,Davis91}. The techniques' results can only give parameter settings that are robust on a particular problem (such as the Traveling Salesman Problem (TSP)), but not on all other problems in a particular domain (such as the combinatorial problem domain where TSP is classified)~\cite{Davis91}. Furthermore, the parameters found in any of these techniques become a liability for GA when the GA structure is modified, such as using another crossover algorithm. Thus, the optimal parameters that resulted from any of the techniques described above may not be good for any GA solving another problem, even to those belonging to the same domain. On the other hand, experimental models can be used to answer the following questions which can not be answered by the techniques used by other researchers:
\begin{enumerate}
 \item Are these genetic operators and their parameters act independently or dependently on GA performance?
 \item If they act independently, how these operators and their parameters affect GA performance? What trend (i.e, linear, quadratic, etc.) these parameters give on GA performance?
 \item If they act dependently, which of these operators and their parameters interactively affect GA performance and how?
\end{enumerate}

Results of past studies~\cite{Pabico96a,Pabico96d} have shown that experimental models can be a standardization technique for GAs. In these studies, an optimal set of genetic operators and parameters for GAs solving problems under the parametric optimization domain was found. The interactive effects of crossover probability, mutation rate, and population size on GA convergence velocity in parameterizing a multiple objective model were determined~\cite{Pabico96d}. The convergence velocity was measured using the offline metric proposed by de~Jong~\cite{DeJong75} while the interaction was measured using a three-factor factorial analysis on the variance of the GA operator-parameter combinations. A GA that uses the combination of 0.60 one-point crossover probability, mutation rate varied over generation and gene representation, and a population density of 30 was found efficient under this problem domain~\cite{Pabico96d}. No explanation, however, was given on how these operators and parameters affect GA performance. In our current effort, we aim to find the same optimal set of genetic operators and parameters for a GA solving problems under the combinatorial optimization domain. In addition, we will attempt to explain how these operators and parameters affect GA performance and investigates whether problem size is also a factor.

In this paper, we report the results of applying experimental models in measuring the interactive effects of operators and parameters on GA performance. Measuring the effects follows that the specific operators and  parameters can be determined to give GA its best performance. Specifically, we used the $n$-factor ANOVA on the interactive effects of operators and parameters to GA convergence. An $n$-factor ANOVA, depending upon a certain probability level, tells how $n$ factors interactively affect a certain response measure (i.e., GA performance) via the goodness-of-fit of the data to the $n$-factor linear model. Although only a few researches have been reported to have used experimental models to compute for and compare different algorithms' performance~\cite{Alviar86,Alviar87,Pabico96a,Pabico96d}, this method offers flexibility and ease of use compared to mathematical analyses or analyses of algorithms. 

Our main objective in this study is to show that experimental models can be a standardization method for GA. Specifically, we aim (1)~to investigate the relationship between the problem size and the GA operators and their parameters, (2)~to investigate whether the selection, crossover and mutation operators act independently on GA performance using $n$-factor ANOVA, and (3)~to suggest genetic operators and their parameters for GA in solving optimization problems under the combinatorial domain. With the promise of GA's general applicability to solve problems, many optimization and search studies can be conducted to try and use this technique. Knowing the relationships between problem size and the genetic operators and parameters that would give GAs an optimal performance, researchers can save time fine tuning their GAs. Further, having known that experimental model can be a standardization technique for GAs, more genetic operators can be devised that can give efficient GAs.

\section{Review of Related Literature}

\subsection[Refinements on Traditional Parameters]{Refinements on Traditional Parameters}
The operators of a traditional GA are selection ($\Omega_s$), crossover ($\Omega_c$), and mutation ($\Omega_m$). The GAs parameter settings are population size ($\lambda$), crossover probability ($p_c$), and mutation rate ($p_m$). A traditional GA uses the roulette wheel selection, one-point crossover with $p_c=0.6$, and bit-mutation with $p_m=0.033$. The population size, set according to the user's discretion, is an important factor because the population of individuals serves as a mechanism with distributed knowledge. This knowledge is being represented by all the genes in the entire population~\cite{Lee94}. Other parameter settings reported in the literature are $p_c=0.6$, $p_m=0.001$, $50\leq\lambda\leq 100$~\cite{DeJong75}, $p_c\in[0.75, 0.95]$, $p_m\in[0.005,0.01]$, $20\leq\lambda\leq 30$~\cite{Schaffer89}, and $p_c=0.95$, $p_m=0.01$, $\lambda=30$~\cite{Grefenstette86}.

GA has been used in parametric optimization and much effort has been put into refining the GA to improve its convergence speed. Researchers~\cite{DeJong75,Grefenstette86,Schaffer89,Davis91} have used four techniques to find good parameter seetings for GA. These techniques are (1)~hand optimization, (2)~using a meta-GA, (3)~brute force search, and (4)~parameters that adapt. de~Jong~\cite{DeJong75} carried out hand optimization to find parameter values for the traditional GA which were good across a set of numerical function optimization problems. The parameter values for single-point crossover and bit mutation were worked out by hand while holding the population size constant. 

Using a meta-GA, the same parameters were optimized by the use of another GA~\cite{Grefenstette86}. With the same set of problems, the GA-optimized GA improved slightly over the GA with hand-optimized parameters. However, a robust parameter setting that would perform well across the range of problems considered was not found. 

Davis~\cite{Davis91} proposed a method that would make the operators evolve or adapt to the problem as the GA iterates. The adapting parameters can be used to study new operators and evaluate its performance. This could be an effective technique for separating the valuable operators from those that are not. Schaffer, et al.~\cite{Schaffer89} sampled the possible parameter settings across a range of values using the same set of problems that Grefenstette~\cite{Grefenstette86} and de Jong~\cite{DeJong75} used. It was concluded that a GA's optimal parameter setting vary from one problem to another. 

\subsection[Measures of GA Performance]{Measures of GA Performance}

de~Jong~\cite{DeJong75} designed two measures to quantify GA's search technique's performance. These are online performance and offline performance. The online performance measures the ongoing performance of the GA and is the running average of all evaluations performed. Mathematically, the online performance is given as
\begin{equation}
  {\rm Online}=\frac{1}{\Lambda}\sum\limits_{i=1}^\Lambda f_i
\end{equation}
where $\Lambda$ is the current number of evaluations and $f_i$ is the $i$th value of the objective function. This measure is appropriate in situations where the cost of evaluating an individual is related in a monotonically increasing way to its fitness value. The offline performance measures convergence and is the running average of the best performance value. The offline performance is computed as
\begin{equation}
  {\rm Offline}=\frac{1}{G}\sum\limits_{i=1}^G f_{{\rm max},i}
\end{equation}
where $G$ is the current generation and $f_{{\rm max},i}={\rm max}\{f_{i,j}:1\leq j\leq\lambda\} $ is the best function value obtained from the $i$th generation. This measure can be used when there is no additional cost for evaluating less-fitted individuals.

\section{Methodology}
\subsection[GA Architectures for TSP]{GA Architectures for TSP}

To solve for TSP, we considered different GA architecture designs. In designing these architectures, the choice for genetic operators is important. Our reasons for choosing the specific genetic operators considered in this study are discussed in the following subsections and are summarized in Table~\ref{Factors}.

\begin{enumerate}
\item {\bf Selection algorithms}. We considered two selection algorithms in this study: Remainder Stochastic Independent Sampling (RSIS) and Stochastic Universal Sampling (SUS). We selected these two algorithms over the usual roullete--wheel method because they are known to have reduced selection bias~\cite{Goldberg89}, giving us assurance that the highly fit individual found at each generation will not be lost by chance in the succeeding generations~\cite{Baker87}. 

\item {\bf Crossover algorithms and probabilities}. We considered two crossover algorithms specifically designed for solving combinatorial problems: Partially Matched Crossover (PMX) and Cycle Crossover (CX). For each algorithm, five crossover probabilities were used, 0.60, 0.65, 0.70, 0.75, and 0.80, which gave us 10 algorithm--probability combinations.

\item {\bf Mutation algorithms}. We decided to use the inversion algorithm to simulate mutation because this method was designed solely for combinatorial problems. We considered  five levels of mutation rates as a parameter for this algorithm: 0.02, 0.04, 0.06, 0.08, and 0.10. 
\end{enumerate}

\begin{table*}[b]
\centering
\caption{Genetic operators and parameters considered in designing a GA for solving TSP.}
\label{Factors}
\begin{tabular}{ c p{0.15cm} c p{0.15cm} c} \hline\hline
{\bf Genetic Operator} &\rule{0.1cm}{0.0in}& {\bf Algorithm} &\rule{0.1cm}{0.0in}& {\bf Parameter Setting}\\ \hline
Selection   & &RSIS        &  &              \\
            & &SUS         &  &              \\
Crossover   & &PMX         &  & 0.60, 0.65, 0.70, 0.75, 0.80  \\
            & &CX          &  & 0.60, 0.65, 0.70, 0.75, 0.80  \\ 
Mutation    & &Inversion   &  & 0.02, 0.04, 0.06, 0.08, 0.10   \\\hline\hline
\end{tabular}
\end{table*}

To determine whether these GA architectures are dependent or independent on the problem size, we considered five different $n$-city TSPs, where $n = \{5, 7, 10, 100, 1000\}$. Varying the size of the problem is important to see whether it will have an effect on the operators and parameters found by ANOVA (i.e., will ANOVA give the same operators and parameters regardless of the size of the problem?). Each $n$-city TSP corresponds to a search space whose size is $n!=\Pi^{n}_{k=1}k = 1 \times 2 \times \cdots \times n$.

We have utilized a total of 100 GA architecures solving TSP under five different problem sizes. We run all GAs until the optimum value for the TSP was reached. For each GA run, we recorded the corresponding offline performance. We performed all GA runs under a multi-programming operating system that is why we only measured the offline performance instead of the actual wall-clock running time.

\subsection[Fitness Function for TSP]{Fitness Function for TSP}
We transformed the TSP into a maximization problem (i.e., the closed-route that will give the maximum profit) and built the problem around a profit matrix, ${\rm \bf PR}$, of known optimum. ${\rm\bf PR}$ is similar to a graph's weighted adjacency matrix, encoding the profit of going from one node to the connecting node. Thus, adjacency and profit between the $i$th and the $j$th nodes is defined if ${\rm \bf PR}_{ij}>0$. If all off-diagonal elements in the matrix are positive, then the graph is fully-connected. In TSP, the value of the elements along the diagonal of the matrix does not matter.

We constructed ${\rm\bf PR}$ creating an $n\times n$ diagonally symmetric positive sparse matrix, ${\rm \bf SMat}$, of random elements and by creating a vector, ${\rm \bf Rt}$, of length $n+1$ whose first $n$~elements are the random permutation of the first $n$~integers and ${\rm\bf Rt}_{n+1}={\rm\bf Rt}_1$. ${\rm \bf Rt}$ is the closed route where the maximum profit can be obtained. For example, if $n=5$, ${\rm \bf SMat}$ and ${\rm \bf Rt}$ might be:
\begin{eqnarray}
   {\rm\bf SMat} &=& \left[\begin{array}{ccccc}
                17 &  22 &  27 &  15 &  17\\
                22 &  16 &  18 &  20 &  15\\
                27 &  18 &  18 &  16 &  17\\
                15 &  20 &  16 &  13 &  16\\
                17 &  15 &  17 &  16 &  10
                           \end{array}\right]\nonumber\\
   {\rm\bf Rt}  &=& \left[\begin{array}{cccccc}
                 4 &   3 &   5 &   1 &   2 & 4\end{array}\right]\label{RouteEg}
\end{eqnarray}
By taking notice of the maximum element of ${\rm\bf SMat}$, $\max({\rm\bf SMat})=27$, and adding it by a constant, say ${\rm MAd}=1$, ${\rm\bf PR}$ can be computed using:
\begin{equation}
   {\rm \bf PR}_{i,j} = \left\{ \begin{array}{ll}
                      {\rm\bf SMat}_{i,j}, & {\rm if\ }i\neq {\rm \bf Rt}_y\\
                                               & {\rm and\ }j \neq {\rm \bf Rt_{y+1}}\\ 
                                               &     \forall 1\leq y\leq n\\
                      {\rm\bf PR}_{j,i}=\max({\rm\bf SMat})+{\rm MAd}, & {\rm otherwise.}
    \end{array}\right.
    \label{Profit}
\end{equation}
The second case, ${\rm\bf PR}_{i,j}={\rm\bf PR}_{j,i}$, in equation \ref{Profit} is necessary so that the same closed route but of different direction (example, in equation~\ref{RouteEg}, ${\rm\bf Rt}^*=[4\quad 2\quad 1\quad 5\quad 3\quad 4]$) will have the same maximum profit. The above equation makes sure that the maximum profit TSP will have a maximum profit of $n\times (\max({\rm\bf SMat})+{\rm MAd})$. With respect to our example, the profit of traversing the optimum route is $5 \times ( 27 + 1) = 168$.

The fitness, $f_i$, of the $i$th randomly generated closed-route can be computed by traversing the route using the profit matrix: 
\begin{equation}
f=\sum_{y=1}^n{\rm\bf PR}_{{\rm\bf Rt}_y,{\rm\bf Rt}_{y+1}}.
\end{equation}
%

\subsection[The Experimental Model]{Experimental Model}
To provide basis for comparison of GA performance as affected by four factors, we used a four-factor ANOVA model. The factors known to have an effect on GA performance are (1)~the algorithm used in simulating selection, (2)~the algorithm and (3)~parameter used in simulating crossover, and (4)~the algorithm and parameter used in simulating mutation. If two selection algorithms produce the same relative GA efficiencies with two crossover and mutation algorithms, then either selection algorithms can be used to evaluate GA efficiencies for any combination of crossover and mutation algorithms. If the results are dependent of selection algorithm, then any one or all combinations of the crossover and mutation algorithms may not be adequate for discriminating among the selection-crossover-mutation algorithm combinations.

The factorial treatment design was used to evaluate whether the four factors act independently on GA performance. The factors that we specifically considered in this study are :
\begin{enumerate}
  \item the selection algorithms ($\Omega_s$) assumed to be discrete with two levels, RSIS and SUS;
  \item the crossover algorithms ($\Omega_c$) assumed to be discrete with two levels, PMX and CX;
  \item the crossover probabilities ($p_c$) assumed to be continuous with five levels from 0.60 to 0.80 on 0.05 intervals; and
  \item the mutation rate ($p_m$) with five continuous levels from 0.02 to 0.10 via 0.02 intervals.
\end{enumerate}
 By determining whether $\Omega_s$, $\Omega_c$, $p_c$, and $p_m$ in combination interact to influence the offline performance of the GA, we can find the combinations of GA operators and parameters that would give the best GA offline performance.

The performance ($P$) of the GA is a function of selection algorithm used ($\Omega_s$), crossover algorithm used ($\Omega_c$), crossover probability used ($p_c$), mutation rate ($p_m$) used, the random error ($\epsilon$\footnote{The random error effect for each test run is assumed to be N($0, \sigma^2$), where N is the normal distribution function with mean 0 and variance $\sigma^2$.}) inherrent to the experiments used which can not be accounted for by $\Omega_s$, $\Omega_c$, $p_c$, and $p_m$, and the interactive effects of $\Omega_s$, $\Omega_c$, $p_c$, and $p_m$. The ANOVA model is therefore
\begin{equation}
\begin{array}{rcl}
   P&=&\epsilon+\alpha_1\Omega_s+\alpha_2\Omega_c + \alpha_3 p_c+\alpha_4 p_m+\\
    & &\quad\alpha_5\Omega_s\Omega_c+\alpha_6\Omega_sp_c+\alpha_7\Omega_s p_m+
            \alpha_8\Omega_cp_c+\\
    & &\quad\alpha_9\Omega_c p_m +\alpha_{10}p_cp_m+\alpha_{11}\Omega_s\Omega_cp_c+\\
    & &\quad\alpha_{12}\Omega_s\Omega_cp_m+
            \alpha_{13}\Omega_sp_cp_m+\alpha_{14}\Omega_cp_cp_m +\\
    & &\quad\alpha_{15}\Omega_s\Omega_cp_cp_m.
\end{array}
\end{equation}

We replicated each GA run four times, each replicate using different random seeds but starting with the same initial population. The analysis of variance tests the hypothesis that $\alpha_i=0,\;\forall\;i$, with a probability of $5\%$.

\subsubsection[Varying the Problem Size]{Varying the Problem Size}
To represent varying problem size, we used different TSP sizes. These sizes are the family of $n$-city TSPs where $n=\{5, 7, 10, 100, 1000\}$. Interestingly, we note here that when solutions are encoded into GA chromosomes using the permutation form, the size of the problem space becomes $n!$. Increasing the search space from $(n-1)!$ is not disadvantageous to GA but rather advantageous because each chromosome can provide $n$ more schemes, a desirable characteristics according to GA's schema theorem~\cite{Goldberg89}. Thus, problem sizes were grouped in terms of the size of the search space brought about by the normal encoding of the solutions to chromosomes. Both $n=7$ and $n=10$ (with search spaces of $6!$ and $9!$, respectively) belong to the intermediate problem size while both $n=100$ and $n=1000$ (with search spaces of $99!$ and $999!$, respectively) belong to the big problem size. $n=5$ represent the small problem size with 120 search points. Because of the extensive computing resources required for performing the experiment involving the bigger problem sizes (i.e, $n=100$ and $n=1000$), only the following levels of genetic parameters were used:
\begin{enumerate}
  \item the crossover probabilities ($p_c$) with three levels 0.60, 0.70, and 0.80 ; and
  \item the mutation rate ($p_m$) with three levels 0.001, 0.010, and 0.100.
\end{enumerate}

\subsubsection[Comparing the Mean GA Performance]{Comparing the Mean GA Performance}
To analyze the factors with continuous levels (i.e., $p_c$ and $p_m$), we partitioned their of sum of squares using trend contrasts. Based on the result of the trend comparison, we performed a regression analysis to model the effect of the factors on GA performance. However, we did not perform the regression when the number of points for regression is less than four. Instead, we performed pairwise comparison on the means of the factors involved. For other factors such as $\Omega_s$ and $\Omega_c$, we conducted a pairwise comparison of means using the Duncan's Multiple range Test (DMRT) at 5\% probability level to explain the  significant effect of these factors to GA performance.

\section{Results and Discussion}

\subsection[Optimum GA Operators for 5-City TSP]{Optimum GA Operators for 5-City TSP}
The ANOVA result for the 5-city TSP shows that there is no $z$-way interaction present, where $z\geq 2$. Table \ref{RESULT05} shows that only $\Omega_c$ has a significant effect on the average GA performance. All other factors have no effect. A simple comparison of means shows that PMX is a better crossover scheme than CX.

\begin{table*}[hbt]
\centering
\vskip 2.0em
\caption{ANOVA table of offline performance of a GA solving a 5--City TSP.}
\label{RESULT05}
\begin{tabular}{l r r r r r} \hline\hline
{\bf Source of} & {\bf Degree of} & {\bf Sum of}  & {\bf Mean}   & $F$-{\bf Value} & Pr$>F$ \\
{\bf Variation} & {\bf Freedom}   & {\bf Squares} & {\bf Square} & \\\hline
Replication             & 3 & 86682.58 & 28894.19 & 1389.71 & 0.0001\\
$\Omega_s$            & 1 &    23.22 &    23.22 &    1.12 & 0.2914\\
$\Omega_c$            & 1 &  1817.18 &  1817.18 &   87.40 & 0.0001\\
$p_c$                 & 4 &    32.72 &     8.18 &    0.39 & 0.8133\\
$p_m$  & 4 &    31.88 &     7.97 &    0.38 & 0.8204\\
$\Omega_s\times\Omega_c$
                       & 1 &     3.97 &     3.97 &    0.19 & 0.6623\\
$\Omega_s\times p_c$   & 4 &    58.95 &    14.73 &    0.71 & 0.5864\\
$\Omega_s\times p_m$   & 4 &   158.94 &    39.73 &    1.91 & 0.1085\\
$\Omega_c\times p_c$   & 4 &    61.31 &    15.32 &    0.74 & 0.5672\\
$\Omega_c\times p_m$   & 4 &    45.87 &    11.46 &    0.55 & 0.6980\\
$p_c\times p_m$        & 16 &    8.19 &     0.51 &    0.02 & 1.0000\\
$\Omega_c\times\Omega_c\times p_c$
                       &  4 &    42.79 &    10.69 &    0.51 & 0.7251\\
$\Omega_s\times\Omega_c\times p_m$
                       & 4 &    28.68 &     7.17 &    0.34 & 0.8475\\
$\Omega_s\times p_c\times p_m$
                       & 16 &    19.81 &     1.23 &    0.06 & 1.0000\\
$\Omega_c\times p_c\times p_m$
                       & 16 &    37.54 &     2.34 &    0.11 & 1.0000\\
$\Omega_s\times\Omega_c\times p_c\times p_m$
                       & 16 &    25.81 &     1.61 &    0.08 & 1.0000\\
Error                & 297  &   6175.07 & 20.79  \\\hline
Total                & 399  &  95254.58 \\\hline\hline
CV=2.07\\
\end{tabular}
\end{table*}

The difference of mean offline performance between PMX and CX can be explained by how these two crossover algorithms behave for some inputs. Given two strings $C_A$ and $C_B$, $C_A\neq C_B$, that encode the solutions to the 5-city TSP, PMX will always create two new strings $C_A^\prime$ and $C_B^\prime$ where $C_i\neq C_i^\prime$ and $f(C_i)\neq f(C_i^\prime)$. However, in CX, for some $C_A$ and $C_B$, the created strings might be the same as the parents strings, $C_A^\prime=C_B$ and $C_B^\prime=C_A$. This defeats the purpose of creating new solutions by crossing-over the parent strings. Take for instance $C_A=\{6, 2, 0, 3, 4, 7, 9, 1, 8, 5\}$ and $C_B=\{7, 0, 5, 2, 8, 1, 3, 4, 9, 6\}$.  Applying CX on these two solutions gives $C_A^\prime=\{7, 0, 5, 2, 8, 1, 3, 4, 9, 6\}$ and $C_B^\prime=\{6, 2, 0, 3, 4, 7, 9, 1, 8, 5\}$. Inputs of this type make CX unable to create new solutions. Table \ref{CX_PMX} shows the relative performance of PMX over CX in terms of new solutions found for all $\Omega_s$--$p_c$--$p_m$ combinations.

\begin{table*}[hbt]\centering
\caption{Comparison of performance between PMX and CX.}
\label{CX_PMX}
\begin{tabular}{c c c c c c c c c c }\hline\hline
              &       &       &  \multicolumn{3}{c}{\bf PMX}&\rule{0.05in}{0.0in} &\multicolumn{3}{c}{\bf CX} \\\cline{4-6}\cline{8-10}
   $\Omega_s$ & $p_c$ & $p_m$ & Actual & Expected & \%  & & Actual & Expected & \% \\
              &       &       & Count  & Count    &     & & Count  & Count    & \\\hline
   RSIS & 0.6 & 0.001 & 2988 & 2988 & 100 & & 1872 & 2992 & 62.57 \\
   RSIS & 0.6 & 0.010 & 2981 & 2981 & 100 & & 1871 & 3004 & 62.28 \\
   RSIS & 0.6 & 0.100 & 2945 & 2945 & 100 & & 1895 & 3014 & 62.87 \\
   RSIS & 0.7 & 0.001 & 3520 & 3520 & 100 & & 2264 & 3504 & 64.61 \\
   RSIS & 0.7 & 0.010 & 3499 & 3499 & 100 & & 2158 & 3504 & 61.82 \\
   RSIS & 0.7 & 0.100 & 3508 & 3508 & 100 & & 2202 & 3516 & 62.63 \\
   RSIS & 0.8 & 0.001 & 4000 & 4000 & 100 & & 2481 & 3981 & 62.32 \\
   RSIS & 0.8 & 0.010 & 3992 & 3992 & 100 & & 2495 & 3986 & 62.09 \\
   RSIS & 0.8 & 0.100 & 4001 & 4001 & 100 & & 2583 & 4055 & 63.70 \\
   SUS  & 0.6 & 0.001 & 2962 & 2962 & 100 & & 1842 & 2989 & 61.63 \\
   SUS  & 0.6 & 0.010 & 2955 & 2955 & 100 & & 1827 & 2975 & 61.41 \\
   SUS  & 0.6 & 0.100 & 2975 & 2975 & 100 & & 1908 & 2943 & 64.83 \\
   SUS  & 0.7 & 0.001 & 2497 & 2497 & 100 & & 2143 & 3488 & 61.44 \\
   SUS  & 0.7 & 0.010 & 3490 & 3490 & 100 & & 2138 & 3474 & 61.54 \\
   SUS  & 0.7 & 0.100 & 3463 & 3463 & 100 & & 2240 & 3461 & 64.72 \\
   SUS  & 0.8 & 0.001 & 3957 & 3957 & 100 & & 2472 & 4001 & 61.78 \\
   SUS  & 0.8 & 0.010 & 3955 & 3955 & 100 & & 2468 & 3991 & 61.84 \\
   SUS  & 0.8 & 0.100 & 3985 & 3985 & 100 & & 2555 & 3965 & 64.44 \\\hline\hline
\end{tabular}
\end{table*}

\subsection[Optimum GA Operators for 7-City and 10-City TSPs]{Optimum GA Operators for 7-City and 10-City TSPs}
A $z$-way interaction is present when simple interaction effects of $z-1$ control variables are not the same at different levels of the $z$th control control variable.  As shown in the analysis of variance tables (Tables~\ref{RESULT07} and \ref{RESULT10}) a four-way interaction is not present among $\Omega_s$, $\Omega_c$, $p_c$, and $p_m$. However, a two-way interaction is present between $\Omega_s$, and $\Omega_c$. The offline performance of the GA behave differently at different $\Omega_s$--$\Omega_c$ combinations (averaged across $p_c$ and $p_m$) which means that varying the values of $p_c$ and $p_m$ will not affect the average offline performance of the GA. The DMRT groupings explain these interactions as shown in Table~\ref{DMRT07}. At 7-City TSP, RSIS--CX, RSIS--PMX, and SUS--PMX are not different from each other while SUS--CX and SUS--PMX have the same effect on GA performance. At 10-City TSP, RSIS--PMX, SUS--CX, and SUS--PMX have the same effect on GA performance and are different from RSIS--CX. The effect of replication (i.e, random seed) on mean GA performance is significant at 7-City TSP only. The presence of significant variability among replications at 7-City TSP suggests that the GA offline performance is dependent on the random number used. This confirms the earlier results of experiments conducted by Goldberg, et al.~\cite{Goldberg92} that GA offline performance is dependent also on the initial population used.

\begin{table*}[hbt]\centering
\caption{ANOVA table of offline performance of a GA solving a 7--City TSP.}
\label{RESULT07}
\begin{tabular}{l r r r r r} \hline\hline
{\bf Source of} & {\bf Degree of} & {\bf Sum of}  & {\bf Mean}   & $F$-{\bf Value} & Pr$>F$ \\
{\bf Variation} & {\bf Freedom}   & {\bf Squares} & {\bf Square} & \\\hline
Replication              &  3 &    502.88 &   167.63 &   7.15 & 0.0001\\
$\Omega_s$   &  1 &    443.50 &   443.50 &  18.92 & 0.0001\\
$\Omega_c$   &  1 &    120.51 &   120.51 &   5.14 & 0.0241\\
$p_c$ &  4 &     57.31 &    14.33 &   0.61 & 0.6550\\
$p_m$    &  4 &    198.60 &    49.65 &   2.12 & 0.0786\\
$\Omega_s\times\Omega_c$ &  1 &    256.91 &   256.91 &  10.96 & 0.0010\\
$\Omega_s\times p_c$     &  4 &     99.53 &    24.88 &   1.06 & 0.3759\\
$\Omega_s\times p_m$     &  4 &    123.41 &    30.85 &   1.32 & 0.2640\\
$\Omega_c\times p_c$     &  4 &     66.55 &    16.64 &   0.71 & 0.5859\\
$\Omega_c\times p_m$     &  4 &    122.37 &    30.59 &   1.30 & 0.2682\\
$p_c\times p_m$          & 16 &    179.65 &    11.23 &   0.48 & 0.9562\\
$\Omega_c\times\Omega_c\times p_c$ &  4 &     45.69 &    79.44 &   1.95 & 0.1024\\
$\Omega_s\times\Omega_c\times p_m$ &  4 &     32.94 &   723.85 &   1.41 & 0.2322\\
$\Omega_s\times p_c\times p_m$     & 16 &      8.98 &   314.55 &   0.38 & 0.9857\\
$\Omega_c\times p_c\times p_m$     & 16 &     16.77 &   186.71 &   0.72 & 0.7782\\
$\Omega_s\times\Omega_c\times p_c\times p_m$ & 16 &     13.90 &   106.09 &   0.59 & 0.8888\\
Error                    &297 & 6963.44 &  23.45\\\hline
Corrected Total          &399 & 10083.49 \\\hline\hline
CV=1.54\\
\end{tabular}
\end{table*}

\begin{table*}[hbt]\centering
\caption{ANOVA table of offline performance of a GA solving a 10--City TSP.}
\label{RESULT10}
\begin{tabular}{l r r r r r} \hline\hline
{\bf Source of} & {\bf Degree of} & {\bf Sum of}  & {\bf Mean}   & $F$-{\bf Value} & Pr$>F$ \\
{\bf Variation} & {\bf Freedom}   & {\bf Squares} & {\bf Square} & \\\hline
Replication              &  3 &    243.23 &    81.08 &   1.06 & 0.3683\\
$\Omega_s$   &  1 &   1512.35 &  1512.35 &  19.69 & 0.0001\\
$\Omega_c$   &  1 &   2461.45 &  2461.45 &  32.05 & 0.0001\\
$p_c$ &  4 &    690.55 &   172.64 &   2.25 & 0.0640\\
$p_m$    &  4 &    619.73 &   154.93 &   2.02 & 0.0920\\
$\Omega_s\times\Omega_c$ &  1 &    735.17 &   735.17 &   9.57 & 0.0022\\
$\Omega_s\times p_c$     &  4 &    331.62 &    82.90 &   1.08 & 0.3668\\
$\Omega_s\times p_m$     &  4 &    273.36 &    68.34 &   0.89 & 0.4703\\
$\Omega_c\times p_c$     &  4 &    585.87 &   146.47 &   1.91 & 0.1092\\
$\Omega_c\times p_m$     &  4 &    122.16 &    30.54 &   0.40 & 0.8103\\
$p_c\times p_m$          & 16 &    816.52 &    51.03 &   0.66 & 0.8282\\
$\Omega_c\times\Omega_c\times p_c$ &  4 &     45.25 &    79.44 &   0.59 & 0.6708\\
$\Omega_s\times\Omega_c\times p_m$ &  4 &     59.33 &   723.85 &   0.77 & 0.5438\\
$\Omega_s\times p_c\times p_m$     & 16 &     47.43 &   314.55 &   0.62 & 0.8694\\
$\Omega_c\times p_c\times p_m$     & 16 &     51.30 &   186.71 &   0.67 & 0.8249\\
$\Omega_s\times\Omega_c\times p_c\times p_m$ & 16 &   1576.64 &    1.28 &   0.59 & 0.2065\\
Error                    &297 & 22811.24 &  76.81\\\hline
Corrected Total          &399 & 34778.01 \\\hline\hline
CV=1.91\\
\end{tabular}
\end{table*}

\begin{table*}[hbt]\centering
\caption{DMRT on mean GA performance for 7--City and 10--City TSPs.}
\label{DMRT07}
\begin{tabular}{c c c} \hline\hline
  $\Omega_s$--$\Omega_c$  & \multicolumn{2}{c}{\bf Mean GA Performance} \\ \cline{2-3}
  {\bf Combination} & {\bf 7-City TSP} & {\bf 10-City TSP}\\\hline
   RSIS--CX               & 316.26a    & 453.58b\\
   RSIS--PMX              & 315.76a    & 461.25a\\
   SUS--CX                & 312.55b    & 460.18a\\
   SUS--PMX               & 315.25ab   & 462.43a\\\hline\hline
  \multicolumn{1}{p{4.5cm}}{\rule{0.25in}{0in}} & \multicolumn{1}{p{4.5cm}}{\rule{0.25in}{0in}} & 
                                  \multicolumn{1}{p{4.5cm}}{\rule{0.25in}{0in}}\\
\end{tabular}
\end{table*}

\subsection[ANOVA Result for 100-City and 1000-City TSPs]{ANOVA Result for 100-City and 1000-City TSPs}

Tables~\ref{RESULT100} and~\ref{RESULT1000} show the ANOVA of GA offline performance for 100-city and 1000-city TSP, respectively. As both results show, two three-way interactions, $\Omega_s$--$\Omega_c$--$p_m$ and $\Omega_s$--$p_c$--$p_m$, exhibit significant differences among their factors. 

DMRT explains the significant differences of these factors (Tables~\ref{DMRT100}, \ref{DMRT100b}, \ref{DMRT100}, and~\ref{DMRT1000b}). Solving a 100-city TSP, the least $\Omega_s$--$\Omega_c$--$p_m$ combination for a GA is SUS, CX, and 0.001, respectively. No specific best combination can be recommended as several combinations can be bests as seen by the DMRT groupings (Table~\ref{DMRT100}).  Three different groupings were identified by DMRT for the $\Omega_s$--$p_c$--$p_m$ combinations (Table~\ref{DMRT100b}). The least $\Omega_s$--$\Omega_c$--$p_m$ combination for a GA that solves 1000-city TSP has $\Omega_s={\rm SUS}$, $\Omega_c={\rm CX}$, and $p_m=0.001$ (Table~\ref{DMRT1000}). Two inferior $\Omega_s$--$p_c$--$p_m$ combinations were also identified , SUS--0.70--0.001 and SUS--0.80--0.001 (Table~\ref{DMRT1000b}). All other combinations are better.

\begin{table*}[hbt]\centering
\caption{ANOVA table of offline performance of a GA solving a 100--City TSP.}
\label{RESULT100}
\begin{tabular}{l r r r r r} \hline\hline
{\bf Source of} & {\bf Degree of} & {\bf Sum of}  & {\bf Mean}   & $F$-{\bf Value} & Pr$>F$ \\
{\bf Variation} & {\bf Freedom}   & {\bf Squares} & {\bf Square} & \\\hline
Replication            & 3  &     240337  &   80112   &  14.96  &   0.0001\\
$\Omega_s$ & 1  &      28871  &   28871   &   5.39  &   0.0222\\
$\Omega_c$  & 1  &     175147  &  175147   &  32.70  &   0.0001\\
$p_c$ 
                       & 2  &       5659  &    2829   &   0.53  &   0.5912\\
$p_m$  & 2  &      25907  &   12953   &   2.42  &   0.0940\\
$\Omega_s\times\Omega_c$
                       & 1  &      18249  &   18249   &   3.41  &  0.0677\\
$\Omega_s\times p_c$
                       & 2  &      11559  &    5779   &   1.08  &   0.3437\\
$\Omega_s\times p_m$
                       & 2  &      88359  &   44179   &   8.25  &   0.0005\\
$\Omega_c\times p_c$
                       & 2  &      31973  &   15986   &   2.98  &   0.0549\\
$\Omega_c\times p_m$
                       & 2  &      51259  &   25629   &   4.78  &   0.0103\\
$p_c\times p_m$        & 4  &      15830  &    3957   &   0.74  &   0.5676\\
$\Omega_c\times\Omega_c\times p_c$
                       & 2  &       2684  &    1342   &   0.25  &   0.7788\\
$\Omega_s\times\Omega_c\times p_m$
                       & 2  &      97260  &   48630   &   9.08  &   0.0002\\
$\Omega_s\times p_c\times p_m$
                       & 4  &      67941  &   16985   &   3.17  &   0.0167\\
$\Omega_c\times p_c\times p_m$
                       & 4  &      12972  &    3243   &   0.61  &   0.6596\\
$\Omega_s\times\Omega_c\times p_c\times p_m$
                       & 4  &      37991  &    9497   &   1.77  &  0.1397\\
Error                & 105  &     562436  &    5356\\\hline
Total                & 143  &    1474443\\\hline\hline
CV=2.33\\
\end{tabular}
\end{table*}

\begin{table*}[hbt]\centering
\caption{DMRT of average GA performance at different combinations of $\Omega_s$, $\Omega_c$, and $p_m$ for 100--city TSP (means with the same letter are not significantly different at 5\% level).}
\label{DMRT100}
\begin{tabular}{c c c c}\hline\hline
             & \multicolumn{3}{c}{$p_m$} \\\cline{2-4}
   $\Omega_s$--$\Omega_c$ & 0.001 & 0.010 & 0.100 \\\hline
   RSIS, CX  &  4599.1a-c & 4626.4ab & 4535.2c\\
   RSIS, PMX &  4639.6a   & 4620.6ab & 4643.7a\\
   SUS, CX   &  4438.2d   & 4555.8bc & 4616.3ab\\
   SUS, PMX  &  4638.7a   & 4618.6ab & 4634.9a\\\hline\hline
   \multicolumn{1}{p{1.5in}}{\rule{0.25in}{0in}} & \multicolumn{1}{p{1.25in}}{\rule{0.25in}{0in}} & 
   \multicolumn{1}{p{1.25in}}{\rule{0.25in}{0in}} & \multicolumn{1}{p{1.25in}}{\rule{0.25in}{0in}} \\
\end{tabular}
\end{table*}

\begin{table*}[hbt]\centering
\caption{ANOVA table of offline performance of a GA solving a 1000--City TSP.}
\label{RESULT1000}
\begin{tabular}{l r r r r r} \hline\hline
{\bf Source of} & {\bf Degree of} & {\bf Sum of}  & {\bf Mean}   & $F$-{\bf Value} & Pr$>F$ \\
{\bf Variation} & {\bf Freedom}   & {\bf Squares} & {\bf Square} & \\\hline
Replication            & 3  &     24256820  &   8085606  &   15.15  &   0.0001\\
$\Omega_s$ & 1  &      2799316  &   2799316  &    5.24  &   0.0240\\
$\Omega_c$  & 1  &     17317700  &  17317700  &   32.44  &   0.0001\\
$p_c$ 
                       & 2  &       575886  &    287943  &    0.54  &   0.5847\\
$p_m$  & 2  &      2576564  &   1288282  &    2.41  &   0.0945\\
$\Omega_s\times\Omega_c$
                       & 1  &      1929517  &   1929517  &    3.61  &   0.0600\\
$\Omega_s\times p_c$
                       & 2  &      1039831  &   519915   &    0.97  &   0.3810\\
$\Omega_s\times p_m$
                       & 2  &      8608709  &   4304354  &    8.06  &   0.0006\\
$\Omega_c\times p_c$
                       & 2  &      3224993  &   1612496  &    3.02  &   0.0530\\
$\Omega_c\times p_m$
                       & 2  &      5050079  &   2525039  &    4.73  &   0.0108\\
$p_c\times p_m$        & 4  &      1675328  &    418832  &    0.78  &   0.5377\\
$\Omega_c\times\Omega_c\times p_c$
                       & 2  &       281360  &    140680  &    0.26  &   0.7688\\
$\Omega_s\times\Omega_c\times p_m$
                       & 2  &      9601609  &   4800804  &    8.99  &   0.0002\\
$\Omega_s\times p_c\times p_m$
                       & 4  &      6794845  &   1698711  &    3.18  &   0.0164\\
$\Omega_c\times p_c\times p_m$
                       & 4  &      1407357  &    351839  &    0.66  &   0.6218\\
$\Omega_s\times\Omega_c\times p_c\times p_m$
                       & 4  &      3891617  &    972904  &    1.82  &   0.1300\\
Error                & 105  &     56052384  &    533832\\\hline
Total                & 143  &    147083926\\\hline\hline
CV=2.01\\
\end{tabular}
\end{table*}

\begin{table*}[hbt]\centering
\caption{DMRT of average GA performance at different combinations of $\Omega_s$, $p_c$, and $p_m$ for 100--city TSP (means with the same letter are not significantly different at 5\% level).}
\label{DMRT100b}
\begin{tabular}{c c c c c}\hline\hline
   &  & \multicolumn{3}{c}{$p_m$}\\ \cline{3-5}
   $\Omega_s$ & $p_c$ & 0.001 & 0.010 & 0.100 \\\hline
   RSIS       & 0.60 & 4613.2a-c  & 4689.1a   & 4567.3bc\\
   RSIS       & 0.70 & 4643.4ab   & 4564.4bc  & 4606.7a-c\\
   RSIS       & 0.80 & 4601.8a-c  & 4617.1a-c & 4597.8a-c\\\
   SUS        & 0.60 & 4551.9bc   & 4561.4bc  & 4621.7a-c\\
   SUS        & 0.70 & 4528.7c    & 4623.6a-c & 4648.4ab\\
   SUS        & 0.80 & 4534.8c    & 4576.6bc  & 4606.7a-c\\\hline\hline
   \multicolumn{1}{p{1.0in}}{\rule{0.25in}{0in}} & \multicolumn{1}{p{1.0in}}{\rule{0.25in}{0in}} & 
   \multicolumn{1}{p{1.0in}}{\rule{0.25in}{0in}} & \multicolumn{1}{p{1.0in}}{\rule{0.25in}{0in}} &
   \multicolumn{1}{p{1.0in}}{\rule{0.25in}{0in}} \\
\end{tabular}
\end{table*}

\begin{table*}[hbt]\centering
\caption{DMRT of average GA performance at different combinations of $\Omega_s$, $\Omega_c$, and $p_m$ for 1000--city TSP (means with the same letter are not significantly different at 5\% level).}
\label{DMRT1000}
\begin{tabular}{c c c c}\hline\hline
             & \multicolumn{3}{c}{$p_m$} \\\cline{2-4}
   $\Omega_s$--$\Omega_c$ & 0.001 & 0.010 & 0.100 \\\hline
   RSIS, CX  & 45960.6a-c  & 46185.2ab & 45338.5c\\
   RSIS, PMX & 46348.5a    & 46149.5ab & 46375.2a\\
   SUS, CX   & 44308.1d    & 45517.6bc & 46129.6ab\\
   SUS, PMX  & 46311.4a    & 46143.2ab & 46276.4a\\\hline\hline
   \multicolumn{1}{p{1.5in}}{\rule{0.25in}{0in}} & \multicolumn{1}{p{1.25in}}{\rule{0.25in}{0in}} & 
   \multicolumn{1}{p{1.25in}}{\rule{0.25in}{0in}} & \multicolumn{1}{p{1.25in}}{\rule{0.25in}{0in}} \\
\end{tabular}
\end{table*}

\begin{table*}[hbt]\centering
\caption{DMRT of average GA performance at different combinations of $\Omega_s$, $p_c$, and $p_m$ for 1000-city TSP (means with the same letter are not significantly different at 5\% level).}
\label{DMRT1000b}
\begin{tabular}{c c c c c}\hline\hline
   &  & \multicolumn{3}{c}{$p_m$}\\ \cline{3-5}
   $\Omega_s$ & $p_c$ & 0.001 & 0.010 & 0.100 \\\hline
   RSIS       & 0.60 & 46085.7a-d & 46844.0a   & 45677.5b-d\\
   RSIS       & 0.70 & 46393.8a-c & 45602.0b-d & 46005.5a-d\\
   RSIS       & 0.80 & 45984.2a-d & 46056.1a-d & 45927.5a-d\\
   SUS        & 0.60 & 45438.5cd  & 45574.8b-d & 46751.1a-d\\
   SUS        & 0.70 & 45226.2d   & 46184.9a-d & 46431.5ab\\
   SUS        & 0.80 & 45264.5d   & 45731.5b-d & 46026.4a-d\\\hline\hline
   \multicolumn{1}{p{1.0in}}{\rule{0.25in}{0in}} & \multicolumn{1}{p{1.0in}}{\rule{0.25in}{0in}} & 
   \multicolumn{1}{p{1.0in}}{\rule{0.25in}{0in}} & \multicolumn{1}{p{1.0in}}{\rule{0.25in}{0in}} &
   \multicolumn{1}{p{1.0in}}{\rule{0.25in}{0in}} \\
\end{tabular}
\end{table*}

\section{Summary and Conclusion}

This study aimed to find the interactive effects of different genetic operators and their parameters on GA offline performance using 4-way ANOVA. Several $n$-city TSPs were considered as test beds, where $n=\{5, 7, 10, 100, 1000\}$. Problem size (i.e., search space) was hypothesized to have an effect on the optimum GA operators and parameter settings. 

ANOVA shows that at a smaller problem size (i.e., 5-city TSP), only $\Omega_c$ has a significant effect on GA offline performance. All other operators and parameters do not affect GA offline performance when the problem size is small. This difference was explained by the way the two $\Omega_c$ algorithms behave. It was found out that PMX is better than CX. When the problem size is intermediate (i.e., 7-City and 10-City TSPs), $\Omega_s$ and $\Omega_c$ interact to affect the mean GA performance. No trend as to what $\Omega_s$--$\Omega_c$ combination is best for this problem size can be concluded as DMRT showed different groupings at different problem sizes.

At bigger problem sizes ($n$-city TSPs where $n=\{100, 1000\}$), the $\Omega_s$--$\Omega_c$--$p_m$ and $\Omega_s$--$p_c$--$p_m$ combinations affect the GA offline performance. No specific behavior on the continuous parameters (i.e, $p_m$ and $p_c$) were found by the regression analysis. Instead DMRT explains the significant three-way interaction among the factors ($\Omega_s$, $\Omega_c$, $p_c$, and $p_m$). Table~\ref{SUMM} summarizes the results of this study.

\begin{table*}[hbt]\centering
\caption{Recommended genetic operator and parameter settings for different problem sizes.}
\label{SUMM}
\begin{tabular}{l c p{6.5cm} } \hline\hline
   {\bf Problem Size}& {\bf Significant} & {\bf Best/Worst Setting}\\
                 &  {\bf Factor}   \\\hline
     5-city TSP & $\Omega_c$ & PMX is better than CX\\ 
    \\
     7-City TSP & $\Omega_s$--$\Omega_c$        & RSIS--CX, RSIS--PMX, and SUS--PMX behave the same while SUS--CX and SUS--PMX have the same effect\\
    \\
    10-city TSP & $\Omega_s$--$\Omega_c$        & RSIS--CX is an inferior combination than the other\\
    \\
    100-city TSP& $\Omega_s$--$\Omega_c$--$p_m$ & both SUS--CX--0.001 is worst\\
                & $\Omega_s$--$p_c$--$p_m$      & No recommendation\\
    \\
    1000-city TSP& $\Omega_s$--$\Omega_c$--$p_m$ & SUS--CX--0.001 is worst\\
                & $\Omega_s$--$p_c$--$p_m$      & both SUS--0.70--0.001 and SUS--0.80--0.001 are inferior\\\hline\hline
\end{tabular}
\end{table*}

It is now therefore concluded that at a smaller problem size, only $\Omega_c$ will have a significant effect on GA offline performance. Between the two $\Omega_c$ considered, PMX has a significantly higher mean GA offline performance than that of CX. When the problem size is intermediate, $\Omega_s$ and $\Omega_c$ interact to affect GA performance. No recommendation as to what combination is best can be given as different groupings were found by DMRT at different problem size within the intermediate range. At bigger problem sizes, the combination of $\Omega_s$--$\Omega_c$--$p_m$ and $\Omega_s$--$p_c$--$p_m$ significantly affect the mean GA offline performance. $\Omega_s={\rm SUS}$, $\Omega_c={\rm CX}$, $p_m=0.001$ is a worst setting for a GA that solves 100-city TSP. The combination of $\Omega_s={\rm SUS}$, $\Omega_c={\rm CX}$, $p_m=0.001$ is worst for a GA that solves a 1000-city TSP. Similarly, both $\Omega_s={\rm SUS}$, $p_c=0.70$, $p_m=0.001$ and $\Omega_s={\rm SUS}$, $p_c=0.80$, $p_m=0.001$ combinations are worst for the same problem.

\bibliography{ga-em}
\bibliographystyle{plain}
\end{document}